%% file: paper.tex
\documentclass[11pt]{article}
\usepackage{acl}
\usepackage[T1]{fontenc}
\usepackage{times}
\usepackage{latexsym}
\usepackage{microtype}
\usepackage{booktabs}
\usepackage{multirow}
\usepackage{amsmath}
\usepackage{amssymb}
\usepackage{graphicx}
\usepackage{xcolor}
\usepackage{colortbl}
\usepackage{url}
\usepackage{enumitem}
\usepackage{tikz}
\usepackage{pgfplots}
\usetikzlibrary{arrows.meta,positioning,fit,calc,shapes.geometric}
\pgfplotsset{compat=1.18}
\usepackage[most]{tcolorbox}
\usepackage{tabularx}
\usepackage{float}
\usepackage{needspace}

\definecolor{promptbg}{HTML}{E8F5E9}
\definecolor{promptheader}{HTML}{2E7D32}
\definecolor{promptheadertext}{HTML}{FFFFFF}
\tcbset{
  sagaprompt/.style={
    enhanced,
    colback=promptbg,
    colframe=promptheader,
    fonttitle=\bfseries\small,
    coltitle=promptheadertext,
    attach boxed title to top left={yshift=-2mm, xshift=4mm},
    boxed title style={colback=promptheader, colframe=promptheader,
                       rounded corners, sharp corners=northeast},
    arc=2pt, outer arc=2pt,
    left=4pt, right=4pt, top=6pt, bottom=4pt,
    before skip=6pt, after skip=6pt,
  }
}

\newcommand{\saga}{SAGA}
\newcommand{\ddpo}{$\Delta$-DPO}
\newcommand{\klsft}{KL-SFT}
\newcommand{\psr}{PSR}
\newcommand{\mattr}{MATTR}
\definecolor{tabgray}{gray}{0.94}
\definecolor{sagablue}{RGB}{30,80,160}
\definecolor{sagagreen}{RGB}{30,130,80}
\definecolor{sagared}{RGB}{180,40,40}
\definecolor{tbdcolor}{RGB}{200,0,0}

\title{\saga{}: Score-Weighted Adaptive Generation Alignment\\
       for Low-Resource Nordic Language Models}

\author{
  \mdseries Hoda Fakharzadehjahromy$^1$ \quad Emil Wiman$^1$ \quad Andreas Bueff$^1$ \\
  Hafsteinn Einarsson$^2$ \quad Fredrik Heintz$^1$ \\[2pt]
  $^1$ Linköping University \quad $^2$ University of Iceland
}

\begin{document}
\maketitle

\begin{abstract}
Language models for low-resource languages still make frequent grammatical errors.
Alignment methods such as RLHF and DPO can correct them. However, they require human judgements, which are scarce for languages such as Icelandic.
Automatic rewards offer an alternative, yet models can learn to exploit them without improving (reward hacking).
We present \saga{} (\textbf{S}core-weighted \textbf{A}daptive \textbf{G}eneration \textbf{A}lignment), which builds preference pairs using dependency parsers. Such parsers already exist for many low-resource languages.
\saga{}'s reward combines parse quality with lexical diversity. It keeps only pairs whose reward gap reaches a threshold and trains on them with DPO.
On Danish, Icelandic and Norwegian Bokm\aa{}l, \saga{} raises reward-parser success by 18 to 28 points. It improves all four model families we test, although one Icelandic run is flagged for reward hacking.
An independent parser confirms the gains, and native speakers prefer \saga{} outputs in Danish (69\%) and Icelandic (80\%). Online RL methods exploit the Icelandic reward. On the same base model, \saga{} scores higher on both parsers.
Existing linguistic resources, combined with safeguards against reward hacking, can therefore stand in for human judgements.
This makes grammatical alignment possible for the many languages that have parsers but no such data.
\end{abstract}

\section{Introduction}

Grammatical correctness is a basic requirement for usable text generation.
Yet language models for low-resource languages still make frequent grammatical errors.
Given the Icelandic fragment \textit{``Þingið samþykkti frumvarpið um''} (``The parliament approved the bill on''), a GPT-SW3-1.3B model often continues with the wrong case.
Token-level metrics miss such errors, but native speakers notice them at once.

\begin{figure}[t!]
\centering
\includegraphics[width=\columnwidth,height=8.5cm]{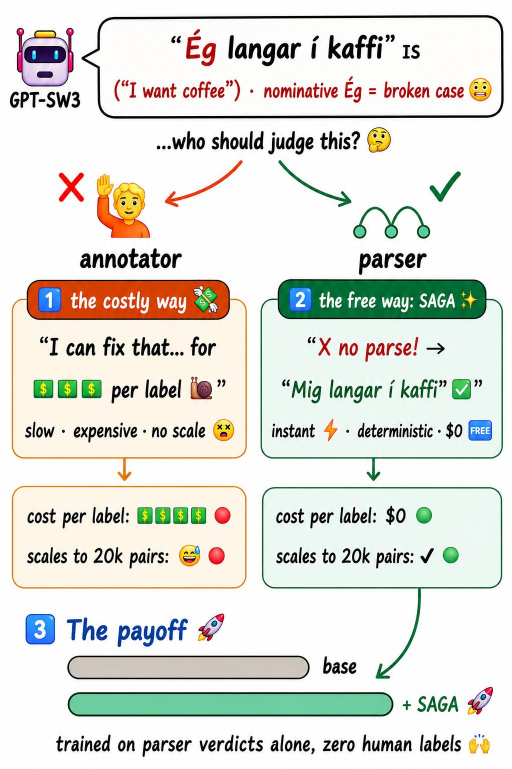}
\caption{A parser-derived preference. The continuation with the wrong case (\textit{Ég}) fails to parse, and the one with the correct case (\textit{Mig}) parses.}
\label{fig:teaser}
\end{figure}

Preference optimisation is one way to correct such behaviour.
Reinforcement learning from human feedback (RLHF) and Direct Preference Optimization (DPO) \cite{rafailov2023dpo} are the standard tools for aligning English models.
Both depend on large collections of human preference judgements, which low-resource languages rarely have.
Qualified annotators are scarce, annotation budgets rarely cover minority languages, and preference datasets of English scale do not exist.
For these languages, preference data rather than model capacity is the main bottleneck.

Many of these languages do have high-quality dependency parsers.
They turn decades of treebank annotation into fast and reproducible judgements about morphology and syntax.
When a continuation with the wrong case fails to parse and the correct one passes (Fig.~\ref{fig:teaser}), the parser's verdict is a preference that costs no annotation.
We ask whether such verdicts can replace human labels in preference optimisation.
LLM judges are the usual alternative, but their verdicts depend on which judge is used (Section~\ref{sec:human}). A parser gives the same verdict every time, and anyone can rerun it. If such a signal can be optimised safely, alignment for the many languages with parsers but no preference data need not wait for annotation.

Using a parser as the reward raises three challenges.
The first is \textbf{reward hacking}.
A model can learn surface patterns that fool the parser while its text gets worse \cite{pang2023reward,gao2023scaling}.
The second is the \textbf{alignment tax}.
Optimising a single reward can make the model forget abilities learned in pretraining \cite{lin2024mitigating}.
The third is \textbf{data scarcity}.
Weak base models produce weak preference candidates, and Icelandic makes up only 2.7\% of the Nordic Pile pretraining corpus \cite{ekgren2023gptSW3}.
The central question is therefore not whether a parser can score text, but whether parser preferences stay reliable once a model is optimised against them. To our knowledge, no prior work tests this for low-resource languages.
\saga{} answers these challenges with a composite reward, DPO on the model's own pairs whose reward gap reaches a threshold (\ddpo{}), and optional regularisation. An ablation tests each part.
Across Danish, Icelandic and Norwegian Bokm\aa{}l and four model families, \saga{} improves grammatical quality without human preference labels. An independent parser and native speakers confirm the gains.

Our contributions are:
\begin{itemize}
  \item \textbf{\saga{}}, a framework that replaces human preference labels with an automatic reward. The reward combines parser quality and lexical diversity, and only pairs whose reward gap reaches a threshold are used for \ddpo{} training.
  \item \textbf{Evidence across three Nordic languages and four model families} that this supervision improves grammatical quality. An independent parser and native speakers confirm the gains, and a Danish model transfers to Swedish with better fluency.
  \item \textbf{A reward-hacking diagnostic and an analysis of failure modes.} The diagnostic flags divergence from an independent parser, lexical collapse and repetition. With it we find that online RL methods game the parser reward on Icelandic, while gap-filtered offline \ddpo{} does not. The best number of passes also depends on the language.
\end{itemize}

\begin{figure*}[t]
\centering
\includegraphics[width=\textwidth]{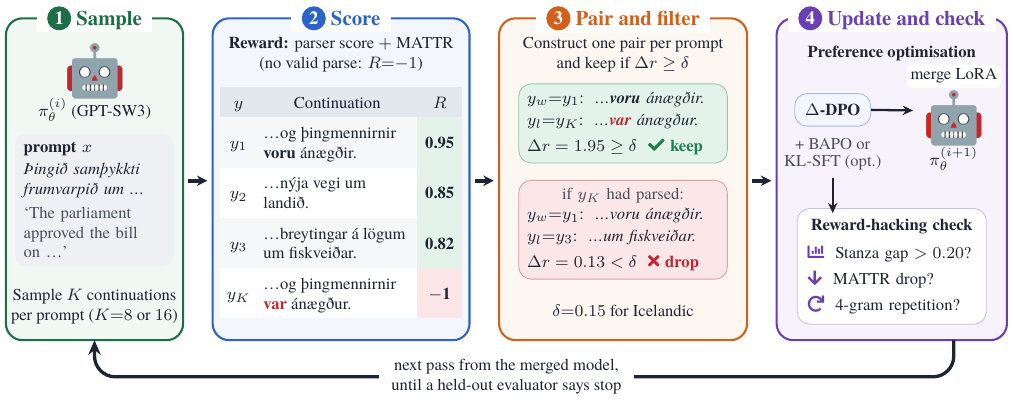}
\caption{The \saga{} pipeline on an Icelandic example. Rewards use Eq.~\ref{eq:reward_is}. The rejected continuation has an agreement error (\textit{var} for \textit{voru}), so $R{=}{-}1$. Settings are in Appendix~\ref{sec:config_appendix}.}
\label{fig:pipeline}
\end{figure*}

\section{Related Work}
\label{sec:background}

\paragraph{RL from verifiable feedback.}
RLHF \cite{christiano2017deep,ouyang2022rlhf} learns from human judgements, but annotation is expensive.
Verifiable rewards, which check an answer against a known solution, give exact signals for math and instruction following \cite{lambert2025tulu}.
Grammar is harder, because a parser checks only one aspect of linguistic quality and a model can exploit it.
Reward hacking \cite{skalse2022reward,gao2023scaling,pang2023reward} and the alignment tax \cite{lin2024mitigating} are the main risks.
However, such rewards exist mainly for tasks with a checkable answer. It is unclear whether a partial verifier such as a parser can be optimised safely.

\paragraph{Preference optimisation.}
DPO \cite{rafailov2023dpo} turns RLHF into a classification loss over chosen and rejected responses, with no explicit reward model.
Its weakness is pair quality.
Pairs whose two responses are of similar quality give a weak signal \cite{lee2026decomposing}. Off-policy pairs from a static dataset also help less than pairs from the current model \cite{lambert2025tulu,croco2026}.
Some variants remove the reference model \cite{meng2024simpo}, and others let the model generate its own rejected responses \cite{yuan2024self}.
Delta learning \cite{geng2025delta} explains why pair quality matters.
Preference tuning can learn from the quality difference between two responses, even when both responses are weak.
OLMo~3 \cite{olmo2026olmo3} uses this at scale by pairing outputs of a stronger and a weaker model. Tülu~3 shows that on-policy completions further improve DPO \cite{lambert2025tulu}.
\citet{lee2026decomposing} separate a generator-level delta, between a strong and a weak model, from a sample-level delta within one pair.
They find that filtering pairs by sample-level delta makes training more data-efficient.
We write \ddpo{} for DPO with delta learning on on-policy pairs \cite{lambert2025tulu,croco2026}.
Our delta is sample-level, the reward gap between two candidates of the same model.
However, all of these methods need a source of preferences, such as human labels, a stronger model or a reward model trained on human preferences.

\paragraph{Parsers and preference supervision.}
CroCo \cite{croco2026} ranks a model's own outputs in many languages with a reward model trained on English preferences, and finds that the gains require on-policy data.
Our cross-lingual warm start is complementary, since it transfers through model weights rather than through a scoring signal.
Multi-objective methods balance several rewards dynamically \cite{yang2024reward}, whereas we use a fixed combination of parse quality and lexical diversity.
However, to our knowledge no prior work uses parser judgements as the main preference signal for low-resource languages, or tests whether they survive optimisation.

\paragraph{Parsing infrastructure and Nordic language resources.}
Universal Dependencies \cite{nivre2020ud} harmonises syntactic annotation across many languages, and Stanza \cite{qi2020stanza} provides parsers trained on it.
We use the Stanza Icelandic model, trained on IcePaHC (the Icelandic Parsed Historical Corpus), as an evaluator that is independent of the Greynir reward parser (Section~\ref{sec:is}).
SpaCy models trained on the Danish UD treebank (DDT) \cite{johannsen2015ddt} and the Norwegian UD treebank \cite{velldal2017universal} serve as reward parsers for Danish and Norwegian.
We also build on Nordic pretrained models \cite{ekgren2023gptSW3,viking2024}, evaluation suites \cite{nielsen2023scandeval,samuel2023norbench} and Icelandic models and corpora \cite{snaebjornsson2022icebert}.
Such parsers exist for many languages that lack preference data, which makes them a practical source of supervision.

\section{The \saga{} Framework}
\label{sec:method}

Standard preference optimisation turns model outputs into training pairs using human judgements.
\saga{} replaces the human with an automatic reward that combines parser quality and lexical diversity.
Pairs whose reward gap is below a threshold are discarded, and the remaining pairs train the model with \ddpo{}.
The diversity term stops the model from trading fluency for grammar. The gap filter removes pairs that small parser errors could flip.

\subsection{Pipeline Overview}
\label{sec:pipeline}

Fig.~\ref{fig:pipeline} shows the pipeline.
In each pass, the model generates $K$ candidates per prompt and scores them with the composite reward (Eq.~\ref{eq:reward}).
The best candidate becomes the chosen response $y_w$. The rejected response $y_l$ is the lower-scoring candidate closest to $\mu{-}2\sigma$ of the $K$ rewards. Pairs whose reward gap $\Delta r{=}R(y_w){-}R(y_l)$ reaches a per-language threshold $\delta$ are kept for \ddpo{} training (Appendix~\ref{sec:config_appendix}).
Some languages also use BAPO (Section~\ref{sec:bapo}) or a KL-SFT warm-up (Table~\ref{tab:config}). KL-SFT is supervised fine-tuning with a KL penalty toward the base model.
We train LoRA adapters (Appendix~\ref{sec:config_appendix}), and the merged checkpoint starts the next pass.

\subsection{Multi-Signal Reward}
\label{sec:reward}

The reward combines three signals for each continuation $g$.
\emph{Parse coverage} $c(g)$ is the fraction of its sentences that parse validly.
\emph{Parse quality} is $d(g)$ for spaCy and $\bar{s}(g)$ for Greynir, defined below.
\emph{Lexical diversity} $m(g)$ is the moving-average type-token ratio (MATTR) of $g$ \cite{covington2010mattr}.
A sentence parses validly when the parser returns a tree for it. With spaCy we also require a single verbal root, a nominal subject and at least five tokens.
A continuation with no valid sentence receives $R(g){=}-1$.
Otherwise, for Danish, Norwegian and Swedish (spaCy),
\begin{equation}
  R(g) = 0.5 + 0.5\,\bigl(0.5\,c(g) + 0.3\,d(g) + 0.2\,m(g)\bigr),
\label{eq:reward}
\end{equation}
where $d(g)$ is the mean dependency depth of the valid sentences, divided by five and capped at one.
For Icelandic (Greynir \cite{thorsteinsson2019greynir}),
\begin{equation}
  R(g) = 0.8\,\bigl(0.5 + 0.5\,c(g)\,\bar{s}(g)\bigr) + 0.2\,m(g),
\label{eq:reward_is}
\end{equation}
where $\bar{s}(g)$ is the mean of the sigmoid $\sigma(s/50)$ over parsed sentences and $s$ is Greynir's parse score.
The constant term rewards any valid parse. The quality terms separate good parses from weak ones. The diversity term penalises repetitive text that a parser alone would accept.
Greynir is grammar-based and rejects case and agreement errors, while the spaCy check targets well-formed clauses.

\subsection{Cross-Lingual Warm-Start}
\label{sec:crosslingual}

Danish and Swedish share verb-second word order and much of their vocabulary.
We therefore start Swedish training from a Danish \saga{} checkpoint, which needs no parallel data and no cross-lingual scoring signal.
Swedish then needs a single \ddpo{} pass on Swedish preference pairs (Section~\ref{sec:sv}).

\subsection{Base-Anchored Policy Optimization}
\label{sec:bapo}

Standard DPO can lower the probability of good continuations, including the chosen response itself \cite{pal2024smaug}.
Following BAPO \cite{bapo2024} and DPO-Positive \cite{pal2024smaug}, we add a one-sided penalty on the chosen response $y_w$ of each pair $(x, y_w, y_l)$ in the preference set $\mathcal{D}$:
\begin{multline}
  \mathcal{L}_\text{BAPO}(\theta) = \mathcal{L}_\text{DPO}(\theta) \\
  + \lambda_B\,\mathbb{E}_\mathcal{D}\bigl[\max\bigl(0,\,
    \log\pi_0(y_w|x) - \log\pi_\theta(y_w|x)\bigr)\bigr]
\label{eq:bapo}
\end{multline}
Here $\mathcal{L}_\text{DPO}$ is the DPO loss, $\pi_\theta$ is the policy being trained, $\pi_0$ is the frozen base model and $\lambda_B$ is the penalty weight.
The penalty is active only when the policy gives the chosen response a lower probability than the base model does.
DPO-Positive places the same hinge inside the DPO sigmoid, and BAPO anchors on a separate base response rather than the chosen one.
The anchor is optional, and $\lambda_B$ is set per language (Table~\ref{tab:config}).

\raggedbottom
\subsection{Automated Reward-Hacking Detection}
\label{sec:hacking}

Before evaluation, we check every trained model for three failure modes.
\textbf{Reward/Stanza divergence.}
The reward/Stanza gap is the fraction of generated sentences that the reward parser accepts, minus the mean Stanza quality $0.6\,c + 0.4\,d$ on the same held-out prompts.
A gap above 0.20 means the model produces text that the reward parser accepts but a parser trained on other data does not. The model is then gaming the reward rather than improving grammar.
\textbf{Lexical collapse.}
A MATTR drop of more than 0.10 from the base model signals repetitive output.
\textbf{Repetition.}
A 4-gram repetition rate (the share of repeated 4-grams in a continuation) above 0.15 confirms such collapse.
We flag a model when any of the three checks fires (Appendix~\ref{sec:config_appendix}). For Icelandic, where domain shift raises every gap, we flag a model whose Reward~PS rises while Stanza~PSR falls below the base.
Flagged models are marked~$\ddagger$ in tables.

\section{Experimental Setup}
\label{sec:setup}

\paragraph{Base models.}
GPT-SW3-1.3B (\texttt{AI-Sweden/\allowbreak gpt-sw3-1b3}) serves as the primary backbone
for all three languages (DA, IS, and NB).
To test whether the method depends on one model family, we repeat the first pass with Viking-13B, Phi-4-14B and, for Norwegian, NorMistral-7B. These differ from GPT-SW3 in pretraining data and architecture (Section~\ref{sec:families}).

\paragraph{Baselines.}
On Icelandic we compare \saga{} with three preference optimisation methods that start from the same base model.
Vanilla DPO trains on static pairs of clean and corrupted sentences, without on-policy generation.
GRPO and sDPO are online methods that sample and score continuations with the same Greynir reward at every training step.
sDPO is score-conditioned DPO, which weights each pair's loss by the chosen response's score.
On Danish we also train with the parser reward alone, without the diversity term.

\paragraph{Training data.}
For DA, we use Danish Wikipedia (20k samples, DDT-filtered, i.e. kept only if the spaCy model trained on DDT scores them above a parse threshold).
For IS, we use the MIM-GOLD corpus \cite{loftsson2010mim} (20k samples, Greynir-filtered, i.e. kept only if Greynir parses them).
For NB, we use HPLT nob\_Latn \cite{de-gibert-etal-2024-new} (20k samples, quality-filtered at $q{\geq}0.65$).
Prompts are the first 4 to 6 tokens of each sentence, and the model generates $K$ continuations per prompt.
Candidates are scored using Equation~\eqref{eq:reward}, and pairs with $\Delta r < \delta$ are discarded.
Evaluation prompts are held out from training ($n{=}200$ for the reward parser and $n{=}400$ for Stanza).

\paragraph{Hyperparameter selection.}
DPO uses $\beta{=}0.1$ \cite{rafailov2023dpo}. All other settings, including learning rate and batch size, are in Appendix~\ref{sec:config_appendix}.
Per-language parameters ($\delta$, $\lambda_B$, iteration count) were selected on held-out prompts, with Stanza~PSR for IS and the reward-parser score for DA and NB. Training stopped when an additional pass no longer improved the validation metric, and these choices were fixed before computing the final reported numbers. For IS, the IcePaHC Stanza evaluation is fully disjoint from training data in both corpus and time period.

\smallskip\noindent\textbf{Evaluation.}
\textit{Parse quality.} Parse Success Rate (\psr{}) is the fraction of generated continuations that contain at least one validly parsed sentence. Validity follows the check of Section~\ref{sec:reward}, which we also apply to Stanza parses. Reward~PS is the PSR of the training parser (Greynir for IS, spaCy for DA and NB). Stanza~PSR uses the Stanza Universal Dependencies parser \cite{qi2020stanza,nivre2020ud}, which has different training data and is never used as a reward.
\textit{Lexical diversity.} We report \mattr{} (window 20 tokens). Perplexity (PPL) is each model's perplexity on 500 held-out Wikipedia sentences, a secondary fluency diagnostic (lower is better).
\textit{Human evaluation.} Native speakers see base and \saga{} continuations of the same prompt in random A/B order and pick the better one or a tie. For Icelandic and Danish they also rate grammaticality and fluency on a 5-point Likert scale. Icelandic uses 43 pairs with 3 speakers per pair (one volunteer and two recruited through Prolific). Danish uses 44 pairs, each rated by 3 of 5 Prolific speakers. Norwegian uses 49 pairs, each rated by 6 Prolific speakers. We report win rates excluding ties with a one-sided pair-level sign test.

\section{Results}
\label{sec:results}

\input{tables/tab_main}

We first ask whether \saga{} improves grammar in each language (Sections~\ref{sec:da}--\ref{sec:nb}). We then ask whether the gains hold across models, languages and tasks (Sections~\ref{sec:families}--\ref{sec:cost}), and whether native speakers agree (Section~\ref{sec:human}). Table~\ref{tab:main} summarises the three languages.

\subsection{Danish}
\label{sec:da}

\paragraph{Parse success rises with every Danish pass.}
One \saga{} pass raises Danish \psr{} from 69\% to \textbf{87.3\%}, and each further pass improves it again. After four passes it reaches \textbf{96.9\%} on average over six seeds (Tab.~\ref{tab:main}).
The independent Stanza parser agrees and rises from 86\% to 99\% (Fig.~\ref{fig:stanza_generalization}). The reward/Stanza gap stays at or below 0.15 at every pass, under the 0.20 threshold.
Lexical diversity does not fall as grammar improves, so the gain does not come from simpler, more repetitive text.
Danish DPO, GRPO and sDPO use the same base and the same reward. They stay at or below 88\% Stanza~\psr{}, and two of them are flagged for hacking (Appendix~\ref{sec:da_baselines}). \saga{} reaches the highest Stanza~\psr{} with the smallest gap.

\subsection{Icelandic}
\label{sec:is}

Icelandic is the hardest test of reward gaming.
Greynir rejects case and agreement errors, but it is also the training reward, so a Greynir gain alone could be gaming.
We therefore rely on three signals the model is not trained on: the independent Stanza parser, native speakers and BÍN morphological recognition.
\saga{} is trained on 20k MIM-GOLD sentences with $\delta{=}0.15$ (Appendix~\ref{sec:is_delta_ablation}).

\paragraph{The Icelandic gains hold on every independent signal.}
Stanza~\psr{} rises from 75.8\% to \textbf{80.3\%} (three-run mean $+$3.3pp), and native speakers prefer \saga{} in \textbf{80\%} of comparisons ($p{<}0.001$). BÍN morphological recognition rises by 0.8pp, which is not significant ($p{=}0.059$).
Greynir Reward~PS rises by 18.3pp, which we treat only as corroborating evidence.
The gap does not grow (0.465 to 0.462), which argues against reward hacking.

\begin{figure}[t]
\centering
\includegraphics[width=\columnwidth]{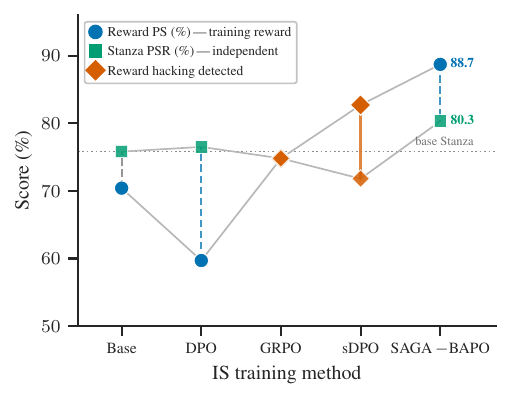}
\caption{Reward~PS (circles, training reward, $n{=}200$) and Stanza~PSR (squares, independent, $n{=}400$) for Icelandic preference optimisation methods with GPT-SW3-1.3B.
The dotted line marks the base Stanza~PSR.
Orange diamonds mark methods whose Reward~PS rises while Stanza~PSR falls below the base.}
\label{fig:is_rl}
\end{figure}

\paragraph{Other preference methods game the Icelandic reward.}
sDPO raises Reward~PS by 12.3pp, but its Stanza~\psr{} falls 4.0pp below the base (Fig.~\ref{fig:is_rl}).
GRPO also ends 1.0pp below the base on Stanza, and vanilla DPO lowers Reward~PS by 10.7pp.
Only \saga{} improves both metrics (Appendix~\ref{sec:is_rl_ablation}).

\paragraph{Icelandic saturates after one pass while Danish keeps improving.}
A second Icelandic pass lowers both Reward~PS and Stanza~\psr{}, while Danish keeps improving through four passes (Tab.~\ref{tab:main}).
Norwegian reaches 94.5\% after a single pass, so we train it for one pass.
More passes do not always help.
Pass~1 without BAPO is the canonical IS configuration. Reward gains alone cannot show when to stop, so a held-out evaluator decides.

\subsection{Norwegian}
\label{sec:nb}

\begin{figure}[t]
\centering
\includegraphics[width=0.85\columnwidth]{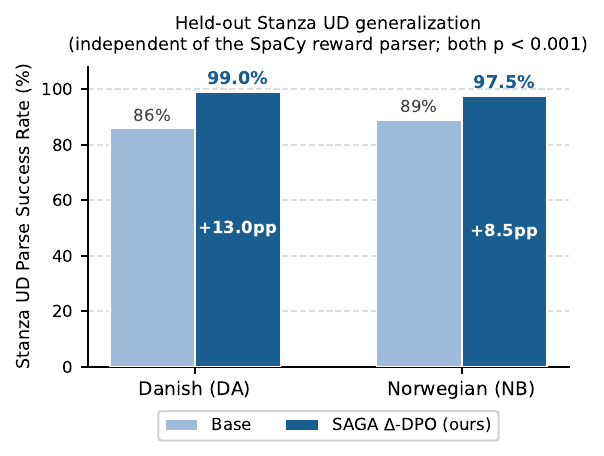}
\caption{Held-out Stanza UD parse success for Danish and Norwegian Bokmål, base model and \saga{}, on treebanks independent of the spaCy reward parser.}
\label{fig:stanza_generalization}
\end{figure}

\paragraph{The gains carry over to Norwegian in a single pass.}
Trained on 20k HPLT sentences, \saga{} raises NB \psr{} from 66.5\% to \textbf{94.5\%} ($p{<}0.001$).
Held-out Stanza~UD confirms the gain (89.0\% to 97.5\%, Fig.~\ref{fig:stanza_generalization}), and the gap stays at 0.156, below the 0.20 threshold.
Lexical diversity rises (Tab.~\ref{tab:main}), and the 4-gram repetition rate falls from 0.022 to 0.003.
On out-of-domain Wikipedia prompts, absolute \psr{} is lower, but the gain is similar (46.9\% to 74.2\%, Tab.~\ref{tab:families}).

\subsection{Model Families}
\label{sec:families}

\input{tables/tab_families}

\paragraph{\saga{} improves every model family we tested.}
We repeat pass~1 with Viking-13B, NorMistral-7B and Phi-4, which differ from GPT-SW3 in pretraining data and architecture (Tab.~\ref{tab:families}).
Every model improves in every language, by 9 to 27 points if we exclude Phi-4 on Icelandic, the only run flagged for hacking.
The smallest model, the Nordic-pretrained GPT-SW3-1.3B, ends highest in all three languages in Table~\ref{tab:families}. Nordic pretraining therefore matters more than scale here (Appendix~\ref{sec:scale_appendix}).

\subsection{Cross-Lingual Transfer to Swedish}
\label{sec:sv}

\paragraph{A Danish checkpoint transfers to Swedish with better fluency.}
Fine-tuned on Swedish Wikipedia, a DA-trained checkpoint reaches 97.0\% \psr{}, not significantly different from training from the base model (97.5\%).
The warm start lowers perplexity from 41.7 to 29.7 ($p{<}0.001$), so parse-aligned initialisation reduces the fluency cost of grammar training (Appendix~\ref{sec:sv_appendix}).

\subsection{Cost and Out-of-Domain Transfer}
\label{sec:cost}

\begin{figure}[t]
\centering
\includegraphics[width=0.92\columnwidth]{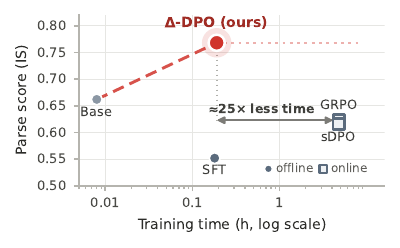}
\caption{Parse score against training time for IS (GPT-SW3-356M) on one A100 GPU. SFT and \ddpo{} times are estimated.}
\label{fig:efficiency}
\end{figure}

\paragraph{\saga{} is cheaper than online RL and transfers to an unseen task.}
\ddpo{} samples all candidates in one offline pass. An Icelandic GPT-SW3-356M run therefore takes an estimated 0.19\,h against a measured 4.8\,h for GRPO and sDPO, and still reaches a higher parse score (Fig.~\ref{fig:efficiency}).
The gains also carry over to Wikipedia summarisation, an unseen task.
There, parse success nearly doubles for Danish and also rises for Icelandic, while plain DPO collapses on Icelandic (Appendix~\ref{sec:summ_transfer}).

\subsection{Pairwise Preference Evaluation}
\label{sec:human}

\input{tables/tab_human}

\paragraph{Native speakers prefer \saga{} in Icelandic and Danish.}
Preference is 80\% for IS and 69\% for DA ($p{<}0.001$, Tab.~\ref{tab:human}). The NB annotators show no preference (52\%, $p{=}0.50$).
Length is unlikely to explain this preference. When \saga{}'s output is not longer, it still wins 77\% (IS) and 76\% (DA) of judgements (Appendix~\ref{sec:length}).
The margin is largest for Icelandic, the most morphologically complex language.
In the Likert ratings, Icelandic raters see the gain mainly as fluency, with almost no change in rated grammaticality ($+$0.01). The Icelandic grammar evidence therefore comes from the parsers.
Danish raters see gains in both (Tab.~\ref{tab:human}), and \saga{} is preferred in all ten Danish annotator subsets. Agreement among the Danish annotators is low (Krippendorff's $\alpha{=}0.12$), so the aggregate preference is more reliable than individual judgements.
For NB the base is already near the grammaticality ceiling (4.7/5 in a separate Likert study of the same base model). The study also used NorMistral-7B, not the model of Table~\ref{tab:main}.
LLM judges are less consistent. On the same 25 IS pairs, Fable-5 (\texttt{claude-fable-5}) prefers \saga{} in 83\% of non-tied cases ($p{<}0.01$). Gemini 2.5 Pro prefers it in only 54\% (not significant, Appendix~\ref{sec:fable_judge_appendix}).

\paragraph{The corrections are the agreement errors the parser checks.}
In one Icelandic pair, the base model continues with \textit{við vorum að fá okkur kaffi og spjallaði} (`we were getting coffee and chatted'). Here the singular verb \textit{spjallaði} does not agree with the plural subject.
\saga{} continues with \textit{við náðum að fá okkur kaffi áður en við fórum} (`we managed to get coffee before we left'), which agrees throughout.
More examples are in Appendix~\ref{sec:qualitative}.

\section{Ablation Study}
\label{sec:ablation}

Table~\ref{tab:ablation} removes one component at a time from the Danish pass-4 model (seed 42, 99.5\% \psr{}). We ablate on Danish because it is the only language trained for four passes.

\input{tables/tab_ablation}

\begin{figure}[t]
\centering
\includegraphics[width=0.92\columnwidth]{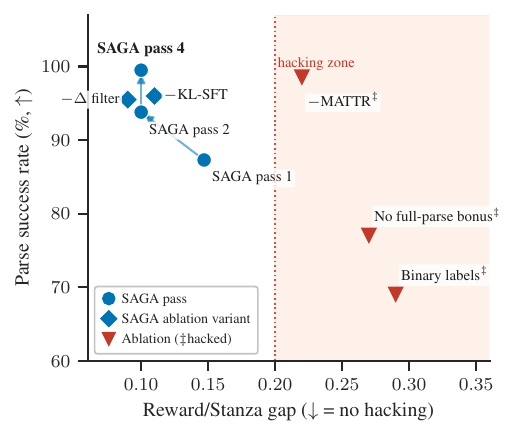}
\caption{Danish \psr{} against the reward/Stanza gap for \saga{} passes and ablation variants. The shaded area marks a gap above 0.20. $^\ddagger$\,hacking detected.}
\label{fig:psr_gap_scatter}
\end{figure}

\paragraph{Both reward signals are needed.}
{
Without the diversity term, \psr{} reaches 98.5\%, but PPL rises from 19.8 to 32.5 and the reward/Stanza gap widens to 0.22, triggering the hacking diagnostic. The model has learned to fool the parser.
With binary 0/1 verdicts instead of graded scores, performance collapses (\psr{}~=~69\%, MATTR 0.66, gap 0.29).
Fig.~\ref{fig:psr_gap_scatter} shows the pattern. Each pass moves to higher parse success and a smaller gap. Every variant that removes a reward safeguard (diversity term, graded scores or full-parse bonus) crosses into the hacking zone.}

\paragraph{The two signals block different failure modes.}
Parse quality alone is not enough.
The policy can raise it with repetitive or overly simple text, and MATTR penalises this.
The structure of the parse reward matters too. A variant that multiplies coverage and depth without the full-parse bonus reaches only 77.0\% \psr{} and enters the hacking region (gap 0.27). The full reward improves grammaticality without crossing the threshold.

\paragraph{The regularisers help only in some languages.}
{\klsft{} is not needed in the first pass, but it stabilises later passes. Without it, pass-4 \psr{} falls by 3.5pp and PPL rises by 7.5.}

{BAPO is language-dependent.
IS performs best without BAPO, where removing it raises both Reward~PS and Stanza~\psr{}. DA and NB benefit from it.
On Danish, removing BAPO raises pass-2 \psr{} but makes fluency much worse (PPL 26.8 vs.\ 19.9, Tab.~\ref{tab:ablation}).
With BAPO and two more passes, pass~4 reaches 99.5\% \psr{} and keeps PPL at 19.8.}

{For NB, every KL-SFT warm-up hurts, and \psr{} falls from 94.5\% to 69.5\% or lower (Tab.~\ref{tab:ablation}).
Lower KL weights hurt more, because weaker regularisation lets the warm-up drift further toward HPLT text, which makes later \ddpo{} training harder.
NB therefore trains without a warm-up ($\lambda{=}0$).}

\paragraph{The gap filter helps on Danish, and on-policy pairs help on Icelandic.}
{\looseness=-1 Removing the gap filter lowers Danish \psr{} by 4pp and raises PPL from 19.8 to 32.7 (Tab.~\ref{tab:ablation}).
On IS, the unfiltered on-policy variant reaches similar Stanza~\psr{} in the fixed-compute sweep (Appendix~\ref{sec:is_delta_ablation}). On IS, the on-policy pairs therefore matter more than the filter.
Static off-policy DPO does worse on IS (Stanza~\psr{} 76.5\% vs.\ 80.3\%) and collapses on IS summarisation transfer (Appendix~\ref{sec:summ_transfer}).}

\paragraph{An imperfect parser is enough.}
A parser misses some errors and rejects some correct sentences (11.7\% for Greynir, Appendix~\ref{sec:is_rl_ablation}), so its verdicts are noisy.
\saga{} does not need them to be exact.
Preference tuning learns from the quality difference within each pair \cite{geng2025delta}, so the parser only has to rank the two outputs correctly more often than not.
The gap filter keeps the pairs where this is most likely, and the diversity term removes the easiest way to please the parser.
The independent parser and native speakers then confirm that the resulting shift toward grammatical text is real.


\section{Conclusion}
\label{sec:conclusion}

{\sloppy\looseness=-1
Parser quality and lexical diversity together give a reliable preference signal.
Across three Nordic languages and four model families, \saga{} improves grammatical quality without human labels, confirmed by an independent parser and native speakers.
The combination matters. Without the diversity signal the model learns to fool the parser, and binary parser verdicts give no gain.
Offline \ddpo{} also avoids the reward gaming of online RL on Icelandic, with an estimated 25 times less training time.
The number of passes should be set per language with a held-out evaluator, not the training reward.
Unlike an LLM judge, a parser gives the same verdict every time, and anyone can rerun it.
Testing \saga{} on other text domains and on non-Nordic languages with parsers but no preference data is the next step.}

\section*{Limitations}
\label{sec:limitations}

\paragraph{Evaluation.}
Danish and Norwegian are trained and partly evaluated with the same spaCy parser family. Stanza evaluation and native speakers reduce, but do not remove, the risk of parser-specific bias.
spaCy often gives an agreement error the same tree as the correct sentence, so for these languages the reward checks clause structure more than agreement.
Parser errors also add noise to the reward, which the gap filter only partly removes.
For Icelandic, Stanza~PSR also guided the choice of $\delta$ and the number of passes. The Stanza model is trained on historical text, so the reward/Stanza gap alone cannot flag Icelandic hacking.
Human evaluation is small (43 to 49 pairs per language). The Norwegian study used a different model from Table~\ref{tab:main}, and the two LLM judges disagree with each other.

\paragraph{Scope.}
\saga{} specialises the model for grammar and lowers Icelandic reading comprehension (Appendix~\ref{sec:nqii}).
We test only Wikipedia, news and web text in three Nordic languages.
Norwegian training and evaluation prompts share a Wikipedia source, but the generated continuations are uncontaminated (Appendix~\ref{sec:contamination_appendix}).
Most results are single runs, except three seeds for Icelandic and six for Danish pass~4.
One pass for the 1.3B model takes about 2 to 4 hours on a single GPU.

\section*{Ethical Considerations}

{\looseness=-1 Our models generate text in Nordic languages and should not be deployed in high-stakes settings without human review.
The underlying base models (GPT-SW3, Viking, and NorMistral) are trained on public web data and may reproduce societal biases. \saga{} introduces no new training data.
Parser rewards encode the standard written language of UD and Greynir. They may penalise dialectal or older usage, as the Greynir/IcePaHC domain gap in Icelandic shows.
Prolific annotators were compensated at the platform-standard rate and all annotators provided informed consent.}

\bibliography{references}

\appendix
\raggedbottom

\section{Per-Language Configuration}
\label{sec:config_appendix}

Table~\ref{tab:config} summarises the reward weights, regularisation, initialisation, and pair-gap threshold used in all reported experiments. The fidelity term is disabled throughout ($\gamma{=}0$).
For \ddpo{}, we use lr~$3{\times}10^{-5}$, batch~32, 2~epochs, $\beta_\text{DPO}{=}0.1$, and repetition penalty~1.3.
For BAPO, we set $\lambda_B{=}0.05$ for DA/NB (fluency preservation) and $\lambda_B{=}0$ for IS/SV.
LoRA adapters use rank $r{=}16$, $\alpha_\text{LoRA}{=}32$ and dropout 0.05 on all linear layers.
Greynir scores are divided by 50 before the sigmoid, so that typical scores fall in its sensitive range.
Each pass samples $K{=}8$ candidates per prompt for DA and $K{=}16$ for IS and NB. The chosen response is the highest-scoring candidate. The rejected response is the lower-scoring candidate closest to $\mu{-}2\sigma$, where $\mu$ and $\sigma$ are the mean and standard deviation of the prompt's candidate scores. This avoids degenerate outliers as rejected responses. MATTR uses a window of 20 tokens.
The reward-hacking check runs on $n{=}200$ held-out prompts. A model is flagged when the reward/Stanza gap exceeds 0.20, MATTR drops by more than 0.10 from the base model, or the 4-gram repetition rate exceeds 0.15. These thresholds were set on the base models.
The reference model is $\pi_0$ (base GPT-SW3-1.3B), and each configuration uses seed~42 unless noted.

\input{tables/tab_config}

\section{Icelandic $\delta$-Threshold Ablation}
\label{sec:is_delta_ablation}

Table~\ref{tab:is_delta} reports a sweep over the \ddpo{} pair-selection threshold $\delta$ for Icelandic (no KL-SFT, single pass on MIM-GOLD). The $\delta{=}0.00^\dagger$ row is a plain DPO baseline. It keeps \saga{}'s on-policy pair generation but removes the gap filter, min-score filter, and BAPO anchor ($\lambda_B{=}0$, min\_chosen\_score~$=$~0). This baseline is distinct from the vanilla DPO setting in Table~\ref{tab:is_rl}, which trains on static clean-vs-corrupted pairs without on-policy generation. All $\delta{>}0$ rows use full \saga{} (BAPO $\lambda_B{=}0.05$, min-score threshold~$=$~0.30). $\delta{=}0.20$ gives the highest Stanza PSR in this fixed-compute sweep (79.8\%). The canonical run of Table~\ref{tab:main} uses $\delta{=}0.15$ without BAPO outside this sweep and reaches 80.3\%. Gap values do not separate the plain-DPO row from the \saga{} rows, because the Icelandic gap reflects the IcePaHC domain shift.

\input{tables/tab_is_delta}

\section{Icelandic RL Method Comparison}
\label{sec:is_rl_ablation}

\input{tables/tab_is_rl_ablation}

Table~\ref{tab:is_rl} compares RL training approaches on IS GPT-SW3-1.3B, evaluated with Greynir Reward~PS (the training reward, independent evaluation in Section~\ref{sec:is}).
Greynir's false-negative rate on 300 gold modern-Icelandic sentences is 11.7\% (verbless parliamentary fragments). The base model's 70.4\% Reward~PS is far below the ${\approx}88\%$ gold-corpus parse ceiling. Most base-model failures therefore reflect genuine grammatical errors rather than coverage gaps.
Vanilla DPO from base (static clean-vs-corrupted preference pairs, no on-policy generation)
\emph{regresses} IS Reward~PS to 59.7\% ($-10.7$pp vs.\ base 70.4\%),
indicating that off-policy preference learning without \saga{}'s on-policy pair
construction and gap filter is harmful. The on-policy no-filter baseline behaves
differently (Table~\ref{tab:is_delta}, $\delta{=}0.00^\dagger$).
\saga{} ($-$BAPO, $\lambda_B{=}0$), starting from the same base, achieves \textbf{88.7\%} Reward~PS ($+18.3$pp),
a $+29.0$pp advantage over vanilla DPO.
The controlled BAPO ablation (bottom two rows in Table~\ref{tab:is_rl}) isolates $\lambda_B$ as the only variable:
$-$BAPO outperforms $+$BAPO by $+$3.6pp Reward~PS (88.7\% vs.\ 85.1\%) and $+$5.0pp Stanza~PSR (80.3\% vs.\ 75.3\%),
suggesting that BAPO over-constrains IS syntactic learning at this $\delta$.

\definecolor{isbg}{HTML}{E3F0FA}
\definecolor{isframe}{HTML}{1565C0}
\definecolor{dabg}{HTML}{FFF3E0}
\definecolor{daframe}{HTML}{E65100}
\definecolor{nbbg}{HTML}{F3E5F5}
\definecolor{nbframe}{HTML}{6A1B9A}
\definecolor{exhdr}{HTML}{FFFFFF}

\tcbset{
  isbox/.style={enhanced, colback=isbg, colframe=isframe,
    fonttitle=\bfseries\small, coltitle=exhdr,
    attach boxed title to top left={yshift=-2mm,xshift=4mm},
    boxed title style={colback=isframe,colframe=isframe,rounded corners,sharp corners=northeast},
    arc=3pt,outer arc=3pt,left=4pt,right=4pt,top=6pt,bottom=4pt,
    before skip=8pt,after skip=8pt},
  dabox/.style={enhanced, colback=dabg, colframe=daframe,
    fonttitle=\bfseries\small, coltitle=exhdr,
    attach boxed title to top left={yshift=-2mm,xshift=4mm},
    boxed title style={colback=daframe,colframe=daframe,rounded corners,sharp corners=northeast},
    arc=3pt,outer arc=3pt,left=4pt,right=4pt,top=6pt,bottom=4pt,
    before skip=8pt,after skip=8pt},
  nbbox/.style={enhanced, colback=nbbg, colframe=nbframe,
    fonttitle=\bfseries\small, coltitle=exhdr,
    attach boxed title to top left={yshift=-2mm,xshift=4mm},
    boxed title style={colback=nbframe,colframe=nbframe,rounded corners,sharp corners=northeast},
    arc=3pt,outer arc=3pt,left=4pt,right=4pt,top=6pt,bottom=4pt,
    before skip=8pt,after skip=8pt},
}

\section{Contamination Analysis}
\label{sec:contamination_appendix}

We measure contamination with 50-character substring windows ($\eta$, \emph{suspicious} $>$0.1, \emph{dirty} $>$0.5).
\textbf{DA:} $\eta{=}0$ and 0\% dirty for both prompts and continuations (uncontaminated).
\textbf{NB:} training and evaluation prompts are both drawn from Norwegian Wikipedia (28\% full-sentence overlap). However, models receive only 2 to 6-word fragments and generate novel continuations ($\eta{<}0.04$ on fragments, 0\% dirty on generated text). Thus, prompt overlap does not inflate PSR.
\textbf{IS:} MIM-GOLD (modern news/web) vs.\ IcePaHC (10th to 18th century historical prose). The corpora are disjoint in genre, register, and period, with empirical $\eta{=}0$.
Contamination therefore does not confound any reported result.

\section{Cross-Domain Stanza UD Generalization}
\label{sec:stanza_generalization}

Fig.~\ref{fig:stanza_generalization} in Section~\ref{sec:nb} reports held-out Stanza UD parse success for Danish and Norwegian Bokmål on independent treebanks that share no overlap with the spaCy reward parser.
\saga{} improves Stanza PSR by $+$13.0pp (DA: 86\%~$\to$~99.0\%) and $+$8.5pp (NB: 89\%~$\to$~97.5\%), both $p{<}0.001$, confirming that gains generalise beyond the parser used during training.

\input{sections/sec_efficiency}

\section{Cross-Lingual Warm Starts}
\label{sec:sv_appendix}

\input{tables/tab_sv}

\begin{figure*}[!ht]
\centering
\includegraphics[width=0.88\textwidth]{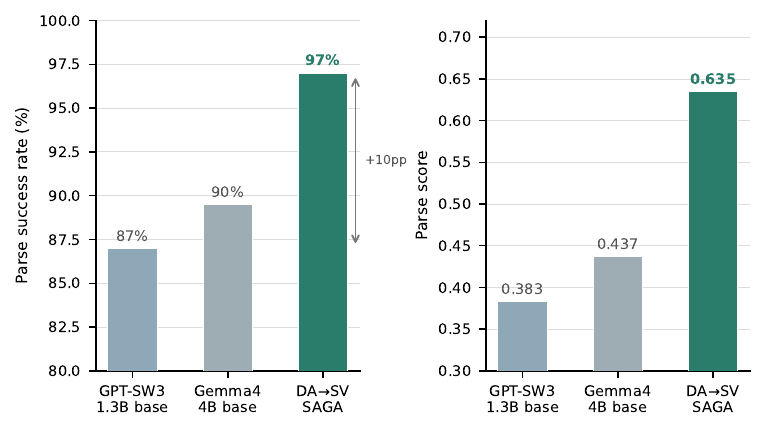}
\caption{Swedish Stanza UD parse success rate (left) and parse score (right) for the GPT-SW3-1.3B and Gemma4-4B base models and the DA$\to$SV \saga{} warm start ($n{=}200$). The warm start is initialised from the Danish iteration-3 checkpoint.}
\label{fig:sv_transfer}
\end{figure*}

\noindent
Table~\ref{tab:sv} and Fig.~\ref{fig:sv_transfer} show the full Swedish transfer results.
A \ddpo{} warm-start trained on DA achieves 97.0\% PSR (PPL~29.7),
which is not significantly different in PSR from a from-base control (97.5\%, $p{=}0.76$), while preserving fluency
significantly better (PPL 29.7 vs.\ 41.7, $p{<}0.001$, paired bootstrap, $n{=}500$).
The from-base control achieves comparable parse success but incurs
substantial fluency regression (PPL 41.7 vs.\ 29.7 for warm-start), consistent with the interpretation that the primary benefit of the DA-to-SV warm-start is
fluency preservation rather than PSR gain.

\input{tables/tab_is_transfer}
\noindent Table~\ref{tab:is_transfer} applies the same idea to Icelandic.
Training from the base model gives the highest Reward~PS (88.7\%). A warm start from the Danish checkpoint reaches 80.9\% with a higher parse-quality score (0.422 vs.\ 0.404).
The warm start therefore trades Reward~PS for parse quality, at a slightly higher PPL (20.3 vs.\ 19.5).

\section{Danish Baselines}
\label{sec:da_baselines}

Table~\ref{tab:da_baselines} evaluates the base model, three preference optimisation baselines and \saga{} with the same scripts, prompts and seed.

\input{tables/tab_da_baselines}

\section{Model Scale and Architecture}
\label{sec:scale_appendix}

\input{tables/tab_nb_scale}
\begin{table*}[!ht]
\begin{minipage}[t]{0.48\textwidth}
\input{tables/tab_da_scale}
\end{minipage}
\hfill
\begin{minipage}[t]{0.48\textwidth}
\input{tables/tab_is_scale}
\end{minipage}
\end{table*}

Table~\ref{tab:families} in the main text compares model families on one shared protocol, with base and trained scores on the same prompts. The tables here report each model on the protocol given in its caption.
Across DA (Table~\ref{tab:da_scale}) and IS (Table~\ref{tab:is_scale}), Nordic pretraining outweighs raw model scale: Phi-4 (14B, multilingual) achieves only 76.8\% DA PSR, whereas GPT-SW3-1.3B (Nordic) reaches 87.3\%. PSR is also non-monotonic within the Nordic family: for DA, GPT-SW3-1.3B (87.3\%) outperforms GPT-SW3-6.7B (84.5\%) and Viking-13B (85.8\%). For IS, GPT-SW3-1.3B achieves the highest Reward PS (88.7\%), outperforming GPT-SW3-6.7B (83.0\%) and Viking-13B (83.1\%), despite Viking-13B's larger base capacity. These results suggest that targeted Nordic pretraining provides stronger grammatical priors than scale alone for morphologically rich Nordic languages.

\section{Summarization Domain Transfer}
\label{sec:summ_transfer}

\begin{figure*}[!ht]
\centering
\includegraphics[width=0.88\textwidth]{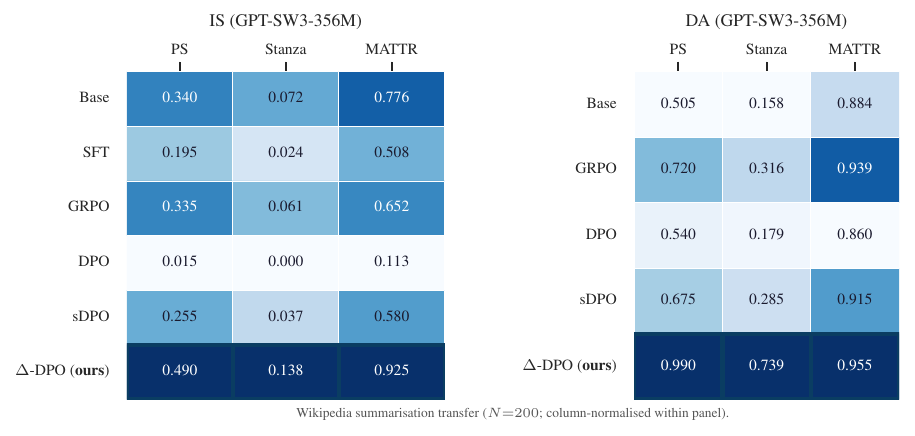}
\caption{Summarization domain-transfer heatmap, IS (left) and DA (right) ($N{=}200$ per language, colour column-normalised within each panel).
None of these models were trained on summarisation.
\ddpo{} obtains the highest parse success (PS), Stanza quality, and lexical diversity (MATTR) in both languages.
DA PS rises from 50.5\% to 99.0\%. IS from 34.0\% to 49.0\%.
Online methods (GRPO, sDPO) transfer substantially less, and plain DPO collapses IS (PS~0.015).}
\label{fig:summ_transfer}
\end{figure*}

Fig.~\ref{fig:summ_transfer} shows the full domain-transfer heatmap.
We evaluate domain transfer by prompting each trained model to perform out-of-domain
Wikipedia article summarization, a task absent from the training distribution.
For IS (GPT-SW3-356M), \ddpo{} achieves PS\,=\,0.490 versus 0.340 for the base model, Stanza\,0.138 versus 0.072,
and MATTR\,0.925 versus 0.776. All baselines (SFT, GRPO, sDPO) fall short, while plain DPO collapses to PS\,0.015.
For DA (GPT-SW3-356M), \ddpo{} achieves PS\,=\,0.990 versus 0.505 for the base model ($+$96\% relative),
with Stanza quality rising from 0.158 to 0.739 and MATTR from 0.884 to 0.955.
Online methods (GRPO: 0.720, sDPO: 0.675) also achieve high DA PS, but with lower quality and diversity than \ddpo{}.
Thus, \ddpo{}'s offline preference signal induces structural grammatical properties that generalise to
unseen domains, whereas online methods (GRPO, sDPO) do not reach the same quality ceiling.

\section{Icelandic Reading Comprehension}
\label{sec:nqii}
We evaluate the base GPT-SW3-1.3B model and the Icelandic \saga{} model of Table~\ref{tab:main} on nqii, the Icelandic reading-comprehension task in EuroEval.
Exact match falls from 24.5 to 14.3, and F1 falls from 46.6 to 38.6.
Grammar-focused training therefore costs factual recall on this task, a trade-off we discuss in the Limitations section.

\section{Length Control for Human Preference}
\label{sec:length}
\input{tables/tab_length}
The composite reward includes dependency depth, so \saga{} could win by writing longer continuations.
Table~\ref{tab:length} uses all pairs annotated to date (48 IS and 42 DA pairs), counting words in the continuation only.
\saga{} outputs are longer on average, but the preference holds when \saga{}'s output is not longer than the base output.
These subsets are small (19 IS and 14 DA pairs), but length is unlikely to explain the preference.

\section{Danish Reward-Hacking Diagnostic}
\label{sec:da_hacking_diagnostic}

Fig.~\ref{fig:psr_gap_scatter} in Section~\ref{sec:ablation} plots PSR against the reward/Stanza gap for all DA ablation configurations.
The shaded zone (gap $>$ 0.20) indicates the reward-hacking threshold. Configurations in this zone are flagged~$\ddagger$ in Table~\ref{tab:ablation}.
Full \saga{} iterations remain in the unshaded region, whereas every ablation that removes a safeguard (parse-only reward, binary labels, no MATTR) enters the hacking zone.

\section{Fable-5 Automated Judge Protocol}
\label{sec:fable_judge_appendix}

\paragraph{Setup.}
As a cross-check on the IS human evaluation (Section~\ref{sec:human}), the Fable-5 judge (\texttt{claude-fable-5}) compared 25 IS pairs (base vs.\ \saga{}~$-$BAPO) from MIM-GOLD prompts. These are separate from the 43 native-speaker pairs.
A/B positions were randomised per pair (seed 42), the temperature was 0.1, and each pair was judged once with the prompt below.

\paragraph{Judge prompt (verbatim).}
{\footnotesize
You are an expert in Icelandic grammar and linguistics. Two language models were given the same Icelandic sentence fragment and asked to complete it. Your task: decide which continuation is more grammatically correct and natural in Icelandic.

\noindent\textbf{Fragment (given to both models):} \textit{\{fragment\}} \textbf{Continuation A:} \textit{\{cont\_a\}} \textbf{Continuation B:} \textit{\{cont\_b\}}

\noindent Evaluate ONLY the continuations; the fragment is fixed input. Assess them on: \textbf{Grammaticality}: correct case (nominative/accusative/dative/genitive), noun-adjective agreement, verb conjugation, and word order. \textbf{Fluency}: naturalness as a native Icelandic speaker would write. \textbf{Coherence}: whether the continuation logically follows from the fragment.

\noindent Then select a winner. If both are equally good or equally bad, choose ``tie''. Respond ONLY with the following JSON (no markdown):
\begin{verbatim}
{ "winner": "A" or "B" or "tie",
  "grammaticality_a": <1-5>,
  "grammaticality_b": <1-5>,
  "fluency_a": <1-5>, "fluency_b": <1-5>,
  "reason": "<key grammatical difference>" }
\end{verbatim}
}

\noindent\textbf{Results.}\quad
\saga{} won 19, tied 2 and lost 4 (83\% excluding ties, $p{<}0.01$, exact binomial). Likert means (Gr./Fl.) were 4.28/2.60 for the base and 4.32/3.44 for \saga{}, labelled $^\S$ in Table~\ref{tab:human}. On the same 25 pairs, Gemini 2.5 Pro gave 13 wins, 1 tie and 11 losses (54\% excluding ties, not significant).

\twocolumn[{%
\section{Qualitative Examples}
\label{sec:qualitative}

Icelandic and Danish examples are drawn from the pairwise evaluation set. Norwegian examples come from the NorMistral-7B outputs used to build the Norwegian pairwise tasks.\par\medskip

\begin{tcolorbox}[isbox, title={Icelandic (IS): GPT-SW3-1.3B base vs.\ \saga{} \ddpo{}}]

\footnotesize
\setlength{\tabcolsep}{4pt}
\renewcommand{\arraystretch}{1.35}
\begin{tabularx}{\linewidth}{p{0.30\linewidth}X X}
\rowcolor{isframe!25}
\textbf{Fragment} & \textbf{Base (GPT-SW3-1.3B)} & \textbf{\saga{} (ours)} \\
\midrule
\emph{Löbbuðum því aftur niðrí bæ þar}
  (\emph{We wandered back downtown where~$\ldots$})
& við vorum að fá okkur kaffi og spjallaði um daginn og veginn.
  (\emph{we were getting coffee and chatted [sg.\ verb, pl.\ subject] about this and that.})
& við náðum að fá okkur kaffi áður en við fórum inn á flugvöll.
  (\emph{we managed to get coffee before heading to the airport.}) \\[4pt]
\emph{Skrifstofa menningarmála hefur eftirlit með framkvæmd stefnumótunar og}
  (\emph{Cultural Affairs Office oversees policy\ldots})
& stefnumótunaráætlunar.
  (\emph{policy-policy planning. [root-word repetition]})
& áætlanagerðar í samræmi við lög og reglugerðir sem settar eru á grundvelli laganna.
  (\emph{planning in accordance with laws and regulations established under the statute.}) \\
\end{tabularx}
\end{tcolorbox}

\begin{tcolorbox}[dabox, title={Danish (DA): GPT-SW3-1.3B base vs.\ \saga{} \ddpo{}}]
\footnotesize
\setlength{\tabcolsep}{4pt}
\renewcommand{\arraystretch}{1.35}
\begin{tabularx}{\linewidth}{p{0.22\linewidth}X X}
\rowcolor{daframe!20}
\textbf{Fragment} & \textbf{Base (GPT-SW3-1.3B)} & \textbf{\saga{} (ours)} \\
\midrule
\emph{I samtlige ministerier prøver man}
  (\emph{In all ministries one tries to~$\ldots$})
& at der skal være et et sammenhængende arbejdsmiljø.
  (\emph{there should be a a [doubled article] coherent work environment.})
& at finde måder, hvorpå man kan undgå, at ansatte oplever det som belastende og krænkende.
  (\emph{find ways to prevent employees from experiencing it as burdensome and demeaning.}) \\[4pt]
\emph{En ung sovjetisk soldat får}
  (\emph{A young Soviet soldier~$\ldots$})
& sitt livs chock när han upptäcker att det inte er sant.
  (\emph{the shock of his life when he discovers it is not true. [code-switched to Swedish]})
& en kniv i ryggen, men han stikker den af for at komme væk fra slagmarken.
  (\emph{a knife in the back, but flees to escape the battlefield.}) \\
\end{tabularx}
\end{tcolorbox}

\begin{tcolorbox}[nbbox, title={Norwegian Bokmål (NB): NorMistral-7B base vs.\ \saga{} \ddpo{} with KL-SFT}]
\footnotesize
\setlength{\tabcolsep}{4pt}
\renewcommand{\arraystretch}{1.35}
\begin{tabularx}{\linewidth}{p{0.22\linewidth}X X}
\rowcolor{nbframe!20}
\textbf{Fragment} & \textbf{Base (NorMistral-7B)} & \textbf{\saga{} (ours)} \\
\midrule
\emph{Lisboa-traktaten, som skulle redde de}
  (\emph{The Lisbon Treaty, which was to save the~$\ldots$})
& europeiske unionen etter den mislykkede Irak krigen og vise at EU er et fredsprosjekt.
  (\emph{European Union [wrong determiner: \textit{de} for \textit{den}] after the failed Irak war [should be one compound] and show EU is a peace project.})
& europeiske sosialdemokratiene fra den nyliberale bølgen og styrke EU i en konkurranseutsatt verden.
  (\emph{European social democracies from the neoliberal wave and strengthen the EU in a competitive world.}) \\[4pt]
\emph{Et tomt blikk var det første moren la}
  (\emph{A blank stare was the first thing the mother noticed~$\ldots$})
& merke til.
  (\emph{notice. \normalfont[minimal; no new information]})
& merke til da hun kom inn i stua etter å ha funnet sønnen livløs.
  (\emph{when she entered the living room after finding her son lifeless.}) \\
\end{tabularx}
\end{tcolorbox}
}]

\end{document}

%% file: tables/tab_main.tex
\begin{table}[t]
\centering\small
\setlength{\tabcolsep}{2.5pt}
\renewcommand{\arraystretch}{1.05}
\caption{Main results with GPT-SW3-1.3B.
Reward~PS: parse success of the training parser (spaCy for DA and NB, Greynir for IS, $n{=}200$).
Stanza~PSR: parse success of the independent Stanza parser (IcePaHC with $n{=}400$ for IS, held-out UD prompts for DA and NB).
Gap: reward/Stanza gap.
$^\dagger$\,$p{<}0.001$ vs.\ base.
$^\ddagger$\,reward hacking detected (Section~\ref{sec:hacking}). --\,=\,not measured.
$^\S$\,mean of six seeds (standard deviation 1.1pp, seed 42: 99.5), gap from seed 42.
For IS, every gap exceeds 0.20 because of the IcePaHC domain shift. Here $^\ddagger$ marks a rise in Reward~PS with a fall in Stanza~PSR below the base.}
\label{tab:main}
\resizebox{\columnwidth}{!}{%
\begin{tabular}{lcccc}
\toprule
\textbf{System} & \textbf{Reward PS} & \textbf{Stanza PSR} & \textbf{MATTR} & \textbf{Gap} \\
\midrule
\multicolumn{5}{l}{\textit{Danish}} \\
Base                       & 69.0 & 86.0 & 0.903 & -- \\
\quad $+$ parse only$^\ddagger$ & 61.0 & -- & 0.831 & ${>}$0.20 \\
\rowcolor{tabgray}
\quad $+$ \saga{} pass~1    & 87.3$^\dagger$ & -- & \textbf{0.909} & 0.147 \\
\quad $+$ \saga{} pass~2    & 93.8$^\dagger$ & \textbf{99.0}$^\dagger$ & 0.906 & 0.100 \\
\quad $+$ \saga{} pass~4    & \textbf{96.9}$^\S$ & -- & -- & 0.10 \\
\midrule
\multicolumn{5}{l}{\textit{Icelandic}} \\
Base                       & 70.4 & 75.8 & 0.840 & 0.465 \\
\quad $+$ DPO              & 59.7 & 76.5 & -- & 0.415 \\
\quad $+$ GRPO$^\ddagger$    & 74.8 & 74.8 & -- & 0.488 \\
\quad $+$ sDPO$^\ddagger$    & 82.7 & 71.8 & -- & 0.491 \\
\rowcolor{tabgray}
\quad $+$ \saga{} pass~1    & \textbf{88.7} & \textbf{80.3} & 0.912 & 0.462 \\
\quad $+$ \saga{} pass~2    & 85.4 & 72.0 & \textbf{0.917} & -- \\
\midrule
\multicolumn{5}{l}{\textit{Norwegian Bokm\aa{}l}} \\
Base                       & 66.5 & 89.0 & 0.900 & -- \\
\rowcolor{tabgray}
\quad $+$ \saga{} pass~1    & \textbf{94.5}$^\dagger$ & \textbf{97.5}$^\dagger$ & \textbf{0.929} & 0.156 \\
\bottomrule
\end{tabular}%
}
\end{table}

%% file: tables/tab_families.tex
\begin{table}[t]
\centering\small
\setlength{\tabcolsep}{2.5pt}
\renewcommand{\arraystretch}{1.05}
\caption{\saga{} pass~1 across model families.
Reward~PS of the base model and after \saga{}, measured on the reward-hacking check prompts, which are the same for both. All rows are matched runs from one training pipeline, so the GPT-SW3 row is not the checkpoint of Table~\ref{tab:main}.
These prompts differ from Table~\ref{tab:main} for DA and NB, so absolute values are lower.
$^\ddagger$\,reward/Stanza gap of 0.563, well above the 0.41 to 0.47 of the other IS models, treated as reward hacking.}
\label{tab:families}
\resizebox{\columnwidth}{!}{%
\begin{tabular}{lccc}
\toprule
\textbf{Model} & \textbf{DA} & \textbf{IS} & \textbf{NB} \\
\midrule
GPT-SW3-1.3B   & 54.5\,$\to$\,\textbf{70.6} & 70.4\,$\to$\,\textbf{88.7} & 46.9\,$\to$\,\textbf{74.2} \\
NorMistral-7B  & --                         & --                         & 55.9\,$\to$\,70.6 \\
Viking-13B     & 57.8\,$\to$\,69.2          & 64.5\,$\to$\,83.1          & 44.0\,$\to$\,69.6 \\
Phi-4-14B    & 59.5\,$\to$\,68.6          & 55.0\,$\to$\,86.6$^\ddagger$ & 46.7\,$\to$\,68.6 \\
\bottomrule
\end{tabular}%
}
\end{table}

%% file: tables/tab_human.tex
\begin{table}[t]
\centering\small
\setlength{\tabcolsep}{3pt}
\renewcommand{\arraystretch}{1.1}
\caption{Pairwise quality evaluation (A/B positions randomised per pair for all evaluations, seed 42).
Win\% excludes ties. W/T/L are judgment-level counts.
IS: 3 annotators/pair. DA: 5 Prolific speakers, 3/pair. NB$^\star$: 6 Prolific speakers/pair, NorMistral-7B with a KL-SFT warm-up, no Likert ratings in this study. IS LLM: 1 judge/pair.
$\Delta$Gr./Fl.\,=\,mean Likert 5-pt grammar/fluency gain (\saga{} vs.\ base).
IS (Fable-5) Likert: rated by the same Fable-5 judge$^\S$.
all other Likert ratings: the native speakers who made the pairwise judgements.
$p$-values: pair-level sign test, one-sided.}
\label{tab:human}
\resizebox{\columnwidth}{!}{%
\begin{tabular}{llccccc}
\toprule
\textbf{Lang.} & \textbf{Comparison} & \textbf{W/T/L} & \textbf{Win\%} & \textbf{$p$} & \textbf{$\Delta$Gr.} & \textbf{$\Delta$Fl.} \\
\midrule
IS (human)              & \saga{} vs.\ base & 83/25/21 & \textbf{80\%} & ${<}0.001$ & $+$0.01           & $+$0.51           \\
IS (Fable-5)            & \saga{} vs.\ base & 19/2/4   & \textbf{83\%} & ${<}0.01$  & $+$0.04$^\S$       & $+$0.84$^\S$       \\
DA (human)              & \saga{} vs.\ base & 59/46/27 & \textbf{69\%} & ${<}0.001$ & $+$0.70            & $+$0.50            \\
NB (human)$^\star$     & \saga{} vs.\ base & 112/80/102 & 52\%         & 0.50       & --                 & --                 \\
\bottomrule
\end{tabular}%
}
\end{table}

%% file: tables/tab_ablation.tex
\begin{table}[t]
\centering\small
\setlength{\tabcolsep}{3pt}
\renewcommand{\arraystretch}{1.1}
\caption{Ablation on Danish (top), KL-SFT weight sensitivity on NB (middle), and fidelity-reward results for IS and NB (bottom).
Full \saga{} pass~4 vs.\ component-removed variants.
$\ddagger$~reward hacking detected (MATTR drop $>$0.10, n-gram repetition $>$0.15, or reward/Stanza gap $>$0.20).
IS rows use Stanza PSR. IS base = 75.8\%. NB rows use spaCy PSR.
For NB, KL-SFT at all $\lambda{>}0$ values \emph{degrades} performance (HPLT corpus distribution is misaligned with the parsing task).
$\lambda{=}0$ (no warm-up) is the NB canonical result.
NB PPL for $\lambda{=}0.10/0.05$ rows were not measured.}
\label{tab:ablation}
\resizebox{\columnwidth}{!}{%
\begin{tabular}{lcccc}
\toprule
\textbf{Variant} & \textbf{PSR} & \textbf{Score} & \textbf{PPL} & \textbf{Gap} \\
\midrule
\rowcolor{tabgray}
Full \saga{} pass~4 & \textbf{99.5} & \textbf{0.747} & 19.8 & 0.10 \\
\quad $-$ \klsft{} ($\lambda{=}0$) & 96.0 & 0.721 & 27.3 & 0.11 \\
\quad $-$ \ddpo{} filter & 95.5 & 0.660 & 32.7 & 0.09 \\
\quad $-$ MATTR (no diversity term)$^\ddagger$ & 98.5 & 0.686 & 32.5 & 0.22 \\
\quad no full-parse bonus$^\ddagger$ & 77.0 & 0.385 & 12.2 & 0.27 \\
\midrule
\multicolumn{5}{l}{\textit{Reward signal comparison (DA, 1.3B)}} \\
\quad Binary-label DPO (0/1 parser verdicts)$^\ddagger$ & 69.0 & 0.382 & 54.3 & 0.29 \\
\midrule
\multicolumn{5}{l}{\textit{BAPO isolation (DA, pass~2)}} \\
\quad \saga{} pass~2 $+$ BAPO ($\lambda_B{=}0.05$) & 93.8 & 0.764 & 19.9 & 0.117 \\
\quad \saga{} pass~2 $-$ BAPO ($\lambda_B{=}0$)    & \textbf{98.3} & \textbf{0.866} & 26.8 & \textbf{0.035} \\
\midrule
\multicolumn{5}{l}{\textit{NB KL-SFT weight sensitivity (HPLT corpus, 1.3B)}} \\
\rowcolor{tabgray}
\quad NB \saga{} (no warm-up, $\lambda{=}0$) & \textbf{94.5} & \textbf{0.645} & 21.8 & 0.156 \\
\quad NB + KL-SFT warm-up ($\lambda{=}0.25$) & 69.5 & 0.403 & 19.2 & 0.248 \\
\quad NB + KL-SFT warm-up ($\lambda{=}0.10$) & 31.4 & 0.258 & — & 0.056 \\
\quad NB + KL-SFT warm-up ($\lambda{=}0.05$) & 28.8 & 0.272 & — & 0.016 \\
\midrule
\multicolumn{5}{l}{\textit{Fidelity reward ablation}} \\
\quad IS + fidelity$^\ddagger$ & 69.0 & 0.358 & 37.2 & 0.43 \\
\quad NB + fidelity & \multicolumn{4}{l}{\textit{0 candidate pairs — no training}} \\
\bottomrule
\end{tabular}%
}
\end{table}

%% file: tables/tab_config.tex
\begin{table}[H]
\centering\small
\setlength{\tabcolsep}{4pt}
\renewcommand{\arraystretch}{1.1}
\caption{Per-language \saga{} hyperparameters.
DA uses separate configurations for pass~1 and later iterations.
DA pass~2 uses $\delta{=}0.15$, and passes 3 and 4 use $\delta{=}0.25$ and keep only pairs whose chosen output scores at least 0.75 and 0.78. SV is initialised from the DA iter-3 checkpoint. Div.\ weight is the effective weight of MATTR in $R$ (Eqs.~\ref{eq:reward} and~\ref{eq:reward_is}).}
\label{tab:config}
\begin{tabular}{lcccc}
\toprule
\textbf{Lang.} & \textbf{Div. weight} & \textbf{KL $\lambda$} & \textbf{$\lambda_B$} & \textbf{$\delta$} \\
\midrule
DA (pass 1) & 0.10 & –    & 0.05 & 0.15 \\
DA (pass 2+)& 0.10 & 0.10 & 0.05 & 0.15/0.25 \\
IS          & 0.20 & –    & 0    & 0.15 \\
NB          & 0.10 & –    & 0.05 & 0.10 \\
SV          & 0.10 & –    & 0    & 0.15 \\
\bottomrule
\end{tabular}
\end{table}

%% file: tables/tab_is_delta.tex
\begin{table}[H]
\centering\small
\setlength{\tabcolsep}{4pt}
\renewcommand{\arraystretch}{1.1}
\caption{Icelandic $\delta$-threshold sweep (GPT-SW3-1.3B, no KL-SFT, fixed compute).
PSR~=~Stanza UD parse success. Reward~=~Greynir parse success. Gap~=~Reward$-$Stanza quality.
$\ddagger$~=~gap~$>$0.20, which for IS reflects the IcePaHC domain shift.
$^\dagger$~=~plain on-policy DPO without the gap filter, min-score filter or BAPO.
All $\delta{>}0$ rows use the gap filter, a min-score filter and BAPO ($\lambda_B{=}0.05$).
Shaded: sweep optimum. \textbf{Bold}: best $\delta{>}0$ row.
The canonical run of Table~\ref{tab:main} uses $\delta{=}0.15$ without BAPO.}
\label{tab:is_delta}
\resizebox{\columnwidth}{!}{%
\begin{tabular}{lcccccc}
\toprule
$\delta$ & \textbf{Reward} & \textbf{PSR} & \textbf{Score} & \textbf{MATTR} & \textbf{Gap} & \textbf{4-gram} \\
\midrule
0.00$^\dagger$ & 85.6 & 80.8 & 0.451 & 0.924 & 0.400$^\ddagger$ & 0.003 \\
\midrule
0.10 & 76.2 & 74.0 & 0.380 & 0.899 & 0.497$^\ddagger$ & 0.011 \\
0.15 & 81.8 & 74.8 & 0.386 & 0.897 & 0.494$^\ddagger$ & 0.015 \\
\rowcolor{tabgray}
0.20 & \textbf{83.5} & \textbf{79.8} & \textbf{0.435} & \textbf{0.918} & \textbf{0.436}$^\ddagger$ & \textbf{0.004} \\
0.25 & 83.7 & 77.5 & 0.399 & 0.886 & 0.525$^\ddagger$ & 0.018 \\
0.40 & 83.1 & 76.0 & 0.394 & 0.922 & 0.429$^\ddagger$ & 0.002 \\
\bottomrule
\end{tabular}%
}
\end{table}

%% file: tables/tab_is_rl_ablation.tex
\begin{table*}[!t]
\centering\small
\setlength{\tabcolsep}{4pt}
\renewcommand{\arraystretch}{1.1}
\caption{Icelandic RL method comparison (GPT-SW3-1.3B).
\textbf{Reward~PS} = Greynir parse success rate ($n{=}200$, same parser used as reward,
held-out prompts disjoint from training).
\textbf{Stanza PSR} = independent Stanza IcePaHC parse success rate ($n{=}400$, archaic text from the 10th to 18th century, not used as a reward).
\textbf{Gap} = Greynir acceptance rate $-$ mean Stanza quality. Every IS gap exceeds 0.20 because of the IcePaHC domain shift, so the gap alone cannot flag IS hacking (base 0.465, \saga{} 0.462).
$^\ddagger$~=~Reward~PS rises while Stanza~PSR falls below the base.
\textbf{Score} = Stanza PSR $\times$ mean Stanza quality.
Bottom two rows: controlled BAPO isolation ($\delta{=}0.15$, all else equal).
Best Reward~PS in \textbf{bold}. \saga{} canonical result shaded.}
\label{tab:is_rl}
\begin{tabular}{llccccc}
\toprule
\textbf{Method} & \textbf{Start} & \textbf{Reward~PS} & \textbf{Stanza PSR} & \textbf{Gap} & \textbf{Score} & \textbf{MATTR} \\
\midrule
Base GPT-SW3-1.3B                                             & —    & 70.4          & 75.8          & 0.465 & 0.364          & 0.840 \\
DPO                                                           & Base & 59.7          & 76.5          & 0.415 & 0.321          & — \\
GRPO$^\ddagger$                                               & Base & 74.8          & 74.8          & 0.488 & 0.384          & — \\
sDPO$^\ddagger$                                               & Base & 82.7          & 71.8          & 0.491 & 0.360          & — \\
\midrule
\saga{} (\ddpo{}, $\delta{=}0.15$, $+$BAPO)                  & Base & 85.1          & 75.3          & 0.466 & 0.409          & — \\
\rowcolor{tabgray}
\saga{} (\ddpo{}, $\delta{=}0.15$, $-$BAPO) \textbf{(ours)}  & Base          & \textbf{88.7}    & \textbf{80.3}    & 0.462 & \textbf{0.404} & \textbf{0.912} \\
\bottomrule
\end{tabular}
\end{table*}

%% file: sections/sec_efficiency.tex
\section{Compute Efficiency of \ddpo{}}
\label{sec:efficiency}

\ddpo{} decouples generation from gradient updates.
All $K{=}16$ completions per prompt are sampled in one offline vLLM pass
(about 0.11\,h for IS 356M), scored CPU-parallel by the dependency parser,
and the DPO loss applied once over surviving pairs (about 0.08\,h). The
estimated total is 0.19\,h.
Online methods (GRPO, sDPO) interleave GPU generation and weight updates
at every training step, requiring 4.8\,h for comparable or lower final parse score.

\begin{figure}[!t]
\centering
\includegraphics[width=\columnwidth]{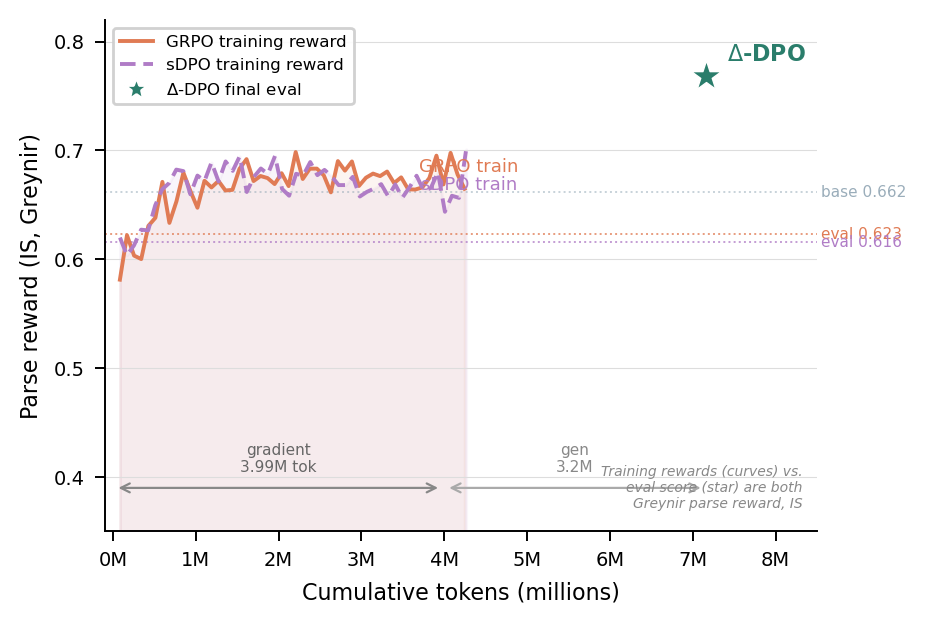}
\caption{Parse reward against cumulative generation tokens for IS (GPT-SW3-356M). The star marks the held-out score of \ddpo{} at its total token budget.}
\label{fig:token_efficiency}
\end{figure}

Figure~\ref{fig:efficiency} in Section~\ref{sec:cost} shows the resulting Pareto frontier:
\ddpo{} achieves parse score 0.768 vs.\ GRPO 0.623 in about $25\times$ less wall-clock time (\ddpo{} and SFT times are estimated, GRPO and sDPO times measured).
Figure~\ref{fig:token_efficiency} complements this with a token view: GRPO and sDPO
plateau near 0.67 training reward at $\sim$4M tokens, yet their held-out parse scores
are only 0.61 to 0.62, indicating reward over-optimisation.
\ddpo{}'s eval score of 0.768 exceeds both training curves without any training--eval gap,
consistent with offline DPO's reduced exposure to greedy reward over-optimisation.

%% file: tables/tab_sv.tex
\begin{table}[H]
\centering\small
\setlength{\tabcolsep}{4pt}
\renewcommand{\arraystretch}{1.1}
\caption{Swedish results. PSR and Score use Stanza UD (\texttt{sv}). PPL on Wikipedia-SV.
The DA$\to$SV warm-start initializes \textbf{GPT-SW3-1.3B} from the DA iter-3 checkpoint.
From-base runs \ddpo{} from the raw GPT-SW3-1.3B checkpoint with SV data only.
Gemma4-4B is an untuned baseline.
Both trained models reach $\approx$97\% PSR. The warm-start achieves this with
substantially lower perplexity (29.7 vs.\ 41.7), indicating better fluency preservation.}
\label{tab:sv}
\begin{tabular}{lccc}
\toprule
\textbf{System} & \textbf{PSR} & \textbf{Score} & \textbf{PPL} \\
\midrule
GPT-SW3-1.3B base         & 87.0 & 0.383 & — \\
Gemma4-4B base            & 89.5 & 0.437 & — \\
\quad From-base \ddpo{}   & 97.5 & 0.654 & 41.7 \\
\rowcolor{tabgray}
\quad DA$\to$SV warm-start (\saga{}) \textbf{(ours)} & \textbf{97.0} & 0.635 & \textbf{29.7} \\
\bottomrule
\end{tabular}
\end{table}

%% file: tables/tab_is_transfer.tex
\begin{table}[t]
\centering\small
\setlength{\tabcolsep}{4pt}
\renewcommand{\arraystretch}{1.1}
\caption{Cross-lingual transfer: DA$\to$IS.
\textbf{Reward~PS} = Greynir parse success rate ($n{=}200$, primary eval).
The DA$\to$IS warm-start initialises \textbf{GPT-SW3-1.3B} from the DA merged checkpoint
and applies IS Greynir reward directly (no KL-SFT, $\delta{=}0.15$).
From-base runs \ddpo{} ($\delta{=}0.15$, $-$BAPO, seed~42) on IS data only.}
\label{tab:is_transfer}
\resizebox{\columnwidth}{!}{%
\begin{tabular}{lccc}
\toprule
\textbf{System} & \textbf{Reward~PS} & \textbf{Score} & \textbf{PPL} \\
\midrule
GPT-SW3-1.3B base                               & 70.4          & 0.364          & 17.0 \\
\quad From-base \ddpo{} ($\delta{=}0.15$, $-$BAPO) & \textbf{88.7} & 0.404          & 19.5 \\
\rowcolor{tabgray}
\quad DA$\to$IS warm-start \textbf{(ours)}       & 80.9          & \textbf{0.422} & 20.3 \\
\bottomrule
\end{tabular}%
}
\end{table}

%% file: tables/tab_da_baselines.tex
\begin{table}[t]
\centering\small
\setlength{\tabcolsep}{4pt}
\renewcommand{\arraystretch}{1.1}
\caption{Danish baselines under one evaluation protocol (GPT-SW3-1.3B, $n{=}200$, seed 42). DPO, GRPO and sDPO are single training runs from the base model with the same Danish Wikipedia data and spaCy reward. \saga{} is pass~4. The spaCy check here is the current one, so absolute values differ from Table~\ref{tab:main}. $^\ddagger$\,reward/Stanza gap $>$0.20.}
\label{tab:da_baselines}
\resizebox{\columnwidth}{!}{%
\begin{tabular}{lcccc}
\toprule
\textbf{System} & \textbf{spaCy PSR} & \textbf{Stanza PSR} & \textbf{Gap} & \textbf{MATTR} \\
\midrule
Base            & 80.0 & 86.5 & -- & 0.899 \\
DPO$^\ddagger$  & 73.5 & 88.0 & 0.226 & 0.899 \\
GRPO$^\ddagger$ & 88.5 & 86.5 & 0.202 & 0.899 \\
sDPO            & 85.0 & 88.0 & 0.175 & 0.902 \\
\rowcolor{tabgray}
\saga{} pass~4  & \textbf{100.0} & \textbf{99.5} & \textbf{0.072} & \textbf{0.913} \\
\bottomrule
\end{tabular}%
}
\end{table}

%% file: tables/tab_nb_scale.tex
%
%
\begin{table}[!t]
\centering\small
\setlength{\tabcolsep}{3pt}
\renewcommand{\arraystretch}{1.1}
\caption{Norwegian \saga{} pass~1 across model architectures.
PSR~=~spaCy \texttt{nb\_core\_news\_lg} reward-parser parse success rate, $n{=}400$,
measured on reward-hacking check prompts (out-of-distribution random Wikipedia
sentences).
This differs from Table~\ref{tab:main}, which uses $n{=}200$ HPLT in-distribution
prompts. Viking-13B scores 98.0\% under that in-distribution protocol.
Nordic~=~Nordic-corpus pretraining. Gap~=~reward/Stanza gap ($\ddagger$~=~$>$0.20).
$\dagger$~=~HPLT nob\_Latn corpus, $\delta{=}0.20$. Others use default $\delta{=}0.10$.
Bottom block: Phi-4 trained on DA, evaluated on NB zero-shot (cross-lingual transfer).}
\label{tab:nb_scale}
\resizebox{\columnwidth}{!}{%
\begin{tabular}{lcccc}
\toprule
\textbf{Model} & \textbf{Params} & \textbf{Nordic} & \textbf{PSR} & \textbf{Gap} \\
\midrule
GPT-SW3          & 1.3B & \checkmark & 74.2          & 0.156 \\
GPT-SW3$^\dagger$& 6.7B & \checkmark & \textbf{75.5} & 0.178 \\
NorMistral       & 7B   & \checkmark & 70.6          & 0.101 \\
Viking \cite{viking2024} & 13B & \checkmark & 69.6          & 0.128 \\
Phi-4            & 14B  & $\times$   & 68.6          & 0.100 \\
\midrule
\multicolumn{5}{l}{\textit{Cross-lingual zero-shot (DA-trained $\to$ NB eval)}} \\
Phi-4 (DA$\to$NB) & 14B & $\times$ & 62.1 & 0.092 \\
\bottomrule
\end{tabular}%
}
\end{table}

%% file: tables/tab_da_scale.tex
\centering\small
\setlength{\tabcolsep}{3pt}
\renewcommand{\arraystretch}{1.1}
\caption{Danish \saga{} pass~1 across model architectures. PSR is spaCy parse success on a separate $n{=}400$ prompt set, except GPT-SW3-1.3B, which is the $n{=}200$ result of Table~\ref{tab:main}. Nordic = Nordic-corpus pretraining. Gap~=~reward/Stanza gap.}
\label{tab:da_scale}
\resizebox{\linewidth}{!}{%
\begin{tabular}{lcccc}
\toprule
\textbf{Model} & \textbf{Params} & \textbf{Nordic} & \textbf{PSR} & \textbf{Gap} \\
\midrule
GPT-SW3       & 1.3B & \checkmark & \textbf{87.3} & 0.147 \\
GPT-SW3       & 6.7B & \checkmark & 84.5          & 0.197 \\
Viking \cite{viking2024} & 13B  & \checkmark & 85.8 & 0.148 \\
Phi-4         & 14B  & $\times$   & 76.8          & 0.145 \\
\bottomrule
\end{tabular}%
}

%% file: tables/tab_is_scale.tex
\centering\footnotesize
\setlength{\tabcolsep}{3pt}
\renewcommand{\arraystretch}{1.1}
\caption{Icelandic \saga{} pass~1 across model scales. Reward~PS = Greynir parse success (reward parser, $n{=}200$). Nordic = Nordic-corpus pretraining. Base rates: 1.3B~70.4\%, Viking-13B~64.5\%, Phi-4~55.0\%. $^\ddagger$ = reward/Stanza gap $>$0.20.}
\label{tab:is_scale}
\resizebox{\linewidth}{!}{%
\begin{tabular}{lccc}
\toprule
\textbf{Model} & \textbf{Params} & \textbf{Nordic} & \textbf{Reward~PS} \\
\midrule
GPT-SW3       & 1.3B & \checkmark & \textbf{88.7}  \\
GPT-SW3       & 6.7B & \checkmark & 83.0           \\
Viking \cite{viking2024} & 13B  & \checkmark & 83.1           \\
Phi-4                    & 14B  & \texttimes & 86.6$^\ddagger$ \\
\bottomrule
\end{tabular}%
}

%% file: tables/tab_length.tex
\begin{table}[H]
\centering\small
\setlength{\tabcolsep}{3pt}
\renewcommand{\arraystretch}{1.1}
\caption{Continuation length and native-speaker preference. Words are counted in the continuation only. Win\% is judgement-level and excludes ties. $p$: one-sided pair-level sign test.}
\label{tab:length}
\resizebox{\columnwidth}{!}{%
\begin{tabular}{llccccc}
\toprule
\textbf{Lang.} & \textbf{Subset} & \textbf{Pairs} & \textbf{Words (\saga{} / base)} & \textbf{W/T/L} & \textbf{Win\%} & \textbf{$p$} \\
\midrule
IS & all pairs            & 48 & 16.4 / 12.8 & 84/36/24 & 78\% & ${<}0.001$ \\
IS & \saga{} not longer   & 19 & --          & 33/14/10 & 77\% & 0.011 \\
IS & \saga{} longer       & 29 & --          & 51/22/14 & 78\% & 0.005 \\
\midrule
DA & all pairs            & 42 & 13.0 / 10.2 & 96/72/42 & 70\% & 0.0025 \\
DA & \saga{} not longer   & 14 & --          & 39/19/12 & 76\% & 0.011 \\
DA & \saga{} longer       & 28 & --          & 57/53/30 & 66\% & 0.054 \\
\bottomrule
\end{tabular}%
}
\end{table}